\documentclass[letterpaper, 10 pt, conference]{ieeeconf}  % Comment this line out if you need a4paper

\IEEEoverridecommandlockouts                              % This command is only needed if 
\usepackage{amsmath,amssymb,amsfonts}
\usepackage[table,xcdraw]{xcolor}
\usepackage{graphicx}
\usepackage{subfigure}
\usepackage{todonotes}
\usepackage{multirow}
\makeatletter
\let\NAT@parse\undefined
\makeatother
\usepackage{makecell} % for \makecell command
\usepackage{booktabs} % for \toprule, \midrule, \bottomrule commands
\usepackage{tikz}
\usepackage{comment}
\usepackage[colorlinks=true, citecolor=black, linkcolor=black]{hyperref}
\usepackage{cite}
\usepackage{algorithm}
\usepackage{algorithmic}

\title{\LARGE \bf
In-Hand Singulation and Scooping Manipulation with \\ a 5 DOF Tactile Gripper
}

\author{Yuhao Zhou$^{1}$, Pokuang Zhou$^{1}$, Shaoxiong Wang$^{2}$ and Yu She$^{1, *}$
\thanks{$^{*}$Corresponding author}
\thanks{$^{1}$School of Industrial Engineering, Purdue University, West Lafayette, IN 47906, USA. {\tt\small \{zhou1437, zhou1458, shey\}@purdue.edu}}%
\thanks{$^{2}$Computer Science and Artificial Intelligence Lab, MIT, Cambridge,
MA 02139, USA. {\tt\small wsxiong13@gmail.com}}%
\thanks{This work was partially supported by USDA under award No. 2023-67021-39072 and No. 2024-67021-42878.}
}

\begin{document}

\maketitle

\begin{abstract}

Manipulation tasks often require a high degree of dexterity, typically necessitating grippers with multiple degrees of freedom (DoF). While a robotic hand equipped with multiple fingers can execute precise and intricate manipulation tasks, the inherent redundancy stemming from its extensive DoF often adds unnecessary complexity. In this paper, we introduce the design of a tactile sensor-equipped gripper with two fingers and five DoF. We present a novel design integrating a GelSight tactile sensor, enhancing sensing capabilities and enabling finer control during specific manipulation tasks. To evaluate the gripper's performance, we conduct experiments involving two challenging tasks: 1) retrieving, singularizing, and classification of various objects embedded in granular media, and 2) executing scooping manipulations of credit cards in confined environments to achieve precise insertion. Our results demonstrate the efficiency of the proposed approach, with a high success rate for singulation and classification tasks, particularly for spherical objects at high as 94.3\%, and a 100\% success rate for scooping and inserting credit cards.

\end{abstract}
%Precise operation including insertion and swiping

% \begin{IEEEkeywords}
% component, formatting, style, styling, insert
% \end{IEEEkeywords}

\section{Introduction}
Dexterous in-hand manipulation stands as a crucial capability for robots tasked with executing precise actions within confined workspaces. Recent advancements in research have showcased the capabilities of robotic hands to emulate precise human-like motions \cite{qi2023hand, yin2023rotating,yuan2023robot}, by leveraging recent strides in visual-tactile sensors, such as GelSight  \cite{yuan2017gelsight}, robots are empowered to perceive object shapes and contact forces during grasping, thereby facilitating autonomous and secure object manipulation. Despite the versatility exhibited by robotic hands with high degrees of freedom, several challenges remain unsolved in the field.

First, methods employing complex finger movements in high DoF often rely on intricate control strategies. Classic model-based approaches necessitate meticulous model and controller designs, often lacking generalization capabilities. Conversely, recent learning-based methods encounter difficulties due to the high dimensionality of dexterous robotic hands, leading to inefficient exploration and the need to tackle the sim-to-real problem. In structured environments where object shapes remain unchanged, a gripper with a simple DoF may suffice to accomplish certain tasks.

Secondly, grappling with small-sized objects in a cluttered environment presents challenges, as multiple fingers may frequently collide with each other, hindering the robot's ability to maintain a balanced grasp on the object. This complexity is further amplified when tasks involve separating objects from cluttered environments, as planning and control issues may arise with the use of multi-finger grippers \cite{wilson2020design}.

\begin{figure}[t]
\vspace{0.2cm}
\centering
\includegraphics[width=0.47\textwidth]{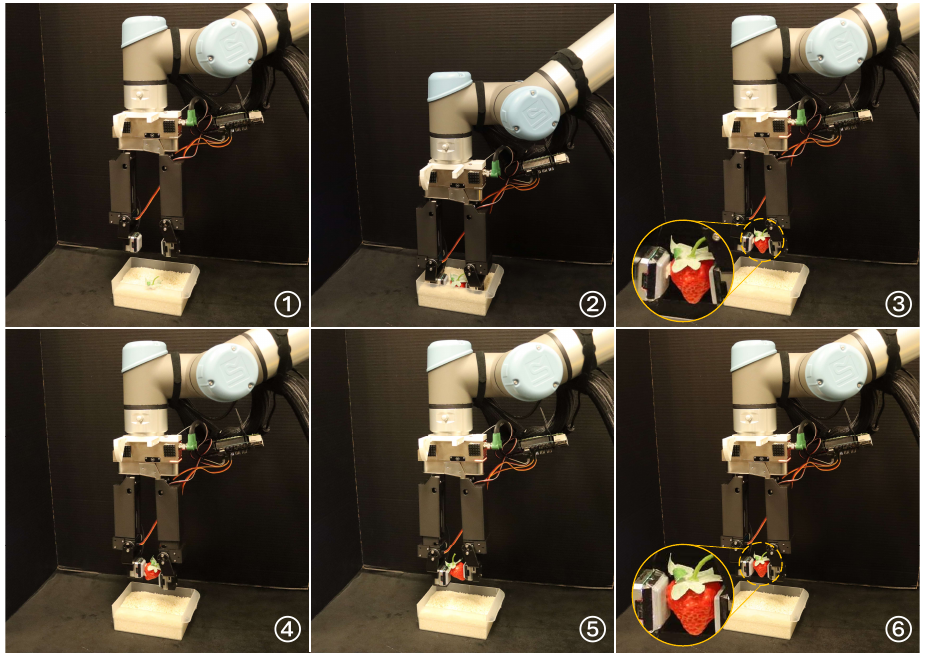}
\caption{In-hand singulation manipulation using the proposed 5 DoF tactile gripper: rubbing off the residual adhesions from the granular media between the finger and the grasped artificial strawberry}
\label{fig:1}
\vspace{-2.0mm}
\end{figure}

Between the common one DoF parallel gripper, which can only accomplish simple grasping tasks, and the multi-finger robotic hands like the 16-DoF Allegro hand, the potential of in-hand manipulation using a fewer DoF gripper remains less explored. In this study, we introduce a novel gripper design based on visual-tactile sensors (Fig. \ref{fig:1}), which enables scooping, in-hand manipulation, precise insertion, and classification tasks using only five degrees of freedom.

The contributions of this study are summarized as follows:

\begin{itemize}
\item We introduce a novel two-finger vision-based tactile gripper with five degrees of freedom. This innovative design facilitates both rolling and linear manipulation, thereby enabling in-hand manipulation tasks.
\item We present a model-based approach that utilizes tactile information to accomplish grasping and singulation of various objects with different properties buried beneath granular media solely through touch sensing.
\item We devise a scooping manipulation maneuver for our gripper capable of handling thin objects, and extending its capabilities to perform precise insertion tasks.
\end{itemize}

% The paper is structured as follows: Section II reviews related works; Section III outlines the problems addressed; Section IV discusses the methodologies for gripper development, perception, object grasping, manipulation, and classification; Section V presents the results and demonstrations, and conclusions and future work are discussed in Section VI.

\section{Related Works}

\subsection{Singulation with Granular Media}

The process of selecting a single object from multiple items within a cluttered environment is commonly referred to as the singulation problem. Previous studies have proposed numerous solutions primarily through vision-based methods \cite{kiatos2019robust,ray2020robotic,zhao2022learning}; however, these approaches cannot be applied to handling small objects buried within granular media. In terms of object identification within granular media, some studies have focused on designing novel tactile sensors. For instance, Patel \emph{et al.} utilized mechanical vibration to fluidize granular media, allowing for deeper penetration to identify different objects buried, using a GelSight sensor \cite{patel2021digger}. Syrymova \emph{et al.} proposed a vibration-based tactile sensor for detecting foreign objects \cite{syrymova2020vibro}. Addressing tactile exploration within granular materials, certain studies have utilized multi-modal sensory data, such as using BioTac to estimate contact states between robotic fingertips and objects within granular media \cite{jia2021tactile, jia2022autonomous}. More recently, researchers have proposed tactile exploration policies aimed at localizing, retrieving, and grasping buried objects using an end-to-end approach with a one DoF parallel gripper \cite{xu2024tactile}. However, existing methods and gripper designs lack the capability to simultaneously grasp objects and complete singulation in a generalized manner. Specifically, none of the previous studies addressed the challenge of grasping objects while removing granular media impurities adhered to their surfaces.

\subsection{Scooping Manipulation}

The grasping or picking of thin objects is a another active research area in robotic manipulation. Particularly challenging is the task of using a two-finger gripper, akin to human capabilities, to pick up plastic cards positioned on a flat supporting surface within constrained environments. Some gripper designs and strategies have been proposed to tackle this problem. For instance, He \emph{et al.} introduced a two-DoF parallel gripper with a variable-length finger, achieving successful grasping through the finger's extending/retracting motion \cite{he2021scooping}. Other studies have adopted similar gripper designs featuring sharp finger tips, demonstrating the effectiveness of a scooping grasping approach \cite{babin2018picking}. Additionally, some two-DoF gripper designs incorporate underactuated and compliant mechanism \cite{odhner2012precision, babin2019stable}. However, due to finger geometry and contact constraints, many previous methods necessitate the motion of the robot arm to assist in the grasping maneuver. In environments where robot joints must be confined within a limited workspace, in-hand manipulation becomes crucial while the robot's end-effector remains fixed. Moreover, integrating a tactile sensor to perform dexterous tasks presents another challenge, particularly regarding the rounded and flat sensor surface lacking a sharp edge for scooping manipulation \cite{do2023densetact}.

\section{Method}
\subsection{Gripper Design}
Prior studies have shown the efficacy of length-adjustable/asymmetric fingers, in tasks such as bin picking \cite{he2021scooping,tong2021dig}, achieved through hand twisting motions to adjust their posture \cite{zuo2021design}. In contrast to other two-finger multi-DoF grippers equipped with vision-based or optical tactile sensors like \cite{wilson2020design, she2021cable, do2023inter}, we aim to achieve greater dexterity in in-hand manipulation by utilizing linear degrees of freedom.

\begin{figure}[t]
\centering
\includegraphics[width=0.5\textwidth]{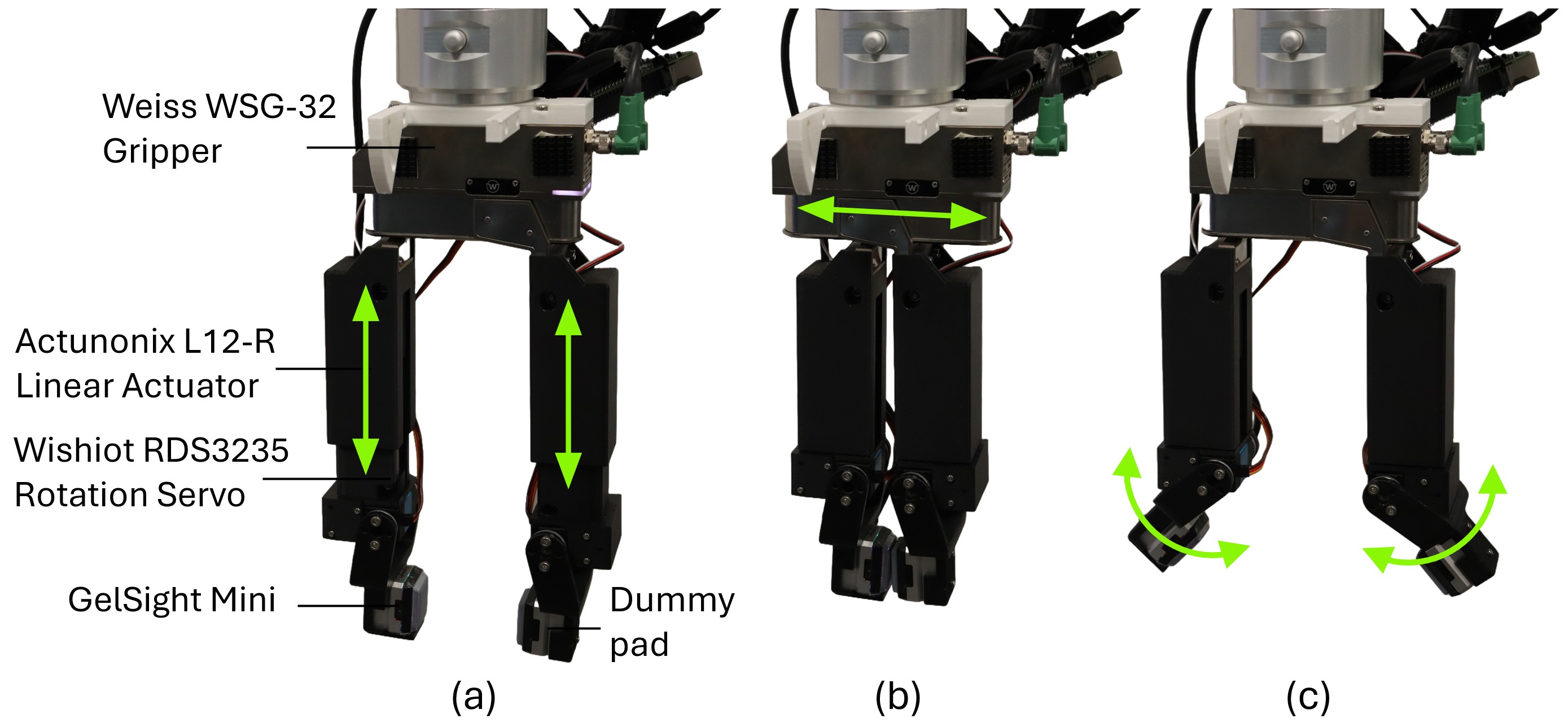}
\vspace{-5.0mm}
\caption{The proposed tactile gripper with five degrees of freedom, composing of a linear actuator and a rotation servo on each finger. One GelSight Mini tactile sensor is mounted one the left fingertip.}
\label{fig:5dof}
% \vspace{-2.0mm}
\end{figure}

The proposed gripper comprises two fingers, each consists of an Actunonix L12-R linear actuator and a Wishiot RDS3235 rotation servo, as depicted in Fig. \ref{fig:5dof}. The linear actuator offers a travel range of $30\mathrm{~mm}$, while the working range of each rotation motor during finger contact is ± $45$°. These fingers are affixed to the parallel Weiss WSG 32 gripper, thereby providing a total of five degrees of freedom. The gripper's arms are fabricated using PLA material via 3D printing, ensuring lightweight construction and reducing the overall load. Our preliminary research on robotic cable manipulation \cite{she2021cable} and cloth manipulation \cite{sunil2023visuotactile} shows that one tactile sensor on the gripper is capable of performing complex manipulation tasks. Therefore, only one GelSight Mini tactile sensor is mounted at the fingertip of the left finger. Additionally, a replaceable flat soft dummy pad is utilized on the other finger to offer compliance, enabling adaptability in handling small objects with soft fingertips.

\subsection{Singulation: Stripping Object from Granular Media}
When humans pick up small shells from the beach, the act of grasping is often followed by a series of delicate in-hand manipulations, involving the fingertips rubbing to remove sand particles that were initially adhered between the object's surface and the fingers. In this section, we outline the approach to address the challenging task of singulation, which involves stripping objects from granular media, mimicking human-like manipulation using the proposed gripper.

\textbf{Stable grasping using tactile sensor:} the first objective of our gripper is to be able to grasp objects within granular material, perform singulation, and subsequently conduct classification to identify the class of the objects. In a tactile-only system, identifying the object becomes extremely challenging if residual granular media remains between the sensor pad surface and the object. During the dynamic rubbing motion of the fingertips, it is essential to ensure adaptability and robustness in grasping. Being too loose may result in dropping the object, while being too tight may lead to singulation failure. To overcome this challenge, we employ the tactile-reactive Model Predictive Control (MPC) controller described in our previous work \cite{xu2024letac}.

Drawing upon the kinematic model and gel properties for a two-finger gripper with one degree of freedom in linear motion, the MPC model based on tactile feedback for grasping, can be formulated as:
\begin{equation}
\label{eq:1}
\left[\begin{array}{c}
c_{n+1} \\
p_{n+1} \\
v_{n+1}
\end{array}\right]=\left[\begin{array}{ccc}
1 & 0 & -K_c \Delta t \\
0 & 1 & \Delta t \\
0 & 0 & 1
\end{array}\right]\left[\begin{array}{c}
c_n \\
p_n \\
v_n
\end{array}\right]+\left[\begin{array}{c}
0 \\
\frac{1}{2} \Delta t^2 \\
\Delta t
\end{array}\right] a_n
\end{equation}
where scalars $c$, $p$, $v$, $a$ refer to the contact area of grasped object, the position, velocity, and the acceleration of the gripper motion. $K_c$ is a scalar factor in the linear assumption of the in the correlation between the contact area size $c$ and the gripping position $p$.

For the cost function, we refined the original method \cite{xu2024letac} in our task to define the feedback vector exclusively based on the contact area from the tactile sensor. This simplifies the problem, considering that the linear and rotational motion of the gripper is confined within a 2D plane and constrained by the gripper's design degrees of freedom settings.

Denote the prediction length as $N$, the cost function can be formulated as follows:
\begin{equation}
\label{eq:2}
J\left(\mathbf{y}_n, \mathbf{a}_n\right)=P \mathbf{e}_{n+N}^T \mathbf{Q} \mathbf{e}_{n+N}+\sum_{k=n}^{n+N-1} \mathbf{e}_k^T \mathbf{Q} \mathbf{e}_k+Q_a a_{k}^2
\end{equation}
where
\begin{equation}
\label{eq:3}
\begin{aligned}
\mathbf{Q} & =\left[\begin{array}{cc}
Q_c & 0 \\
0 & Q_v
\end{array}\right], \\
\end{aligned}
\end{equation}

\begin{equation}
\label{eq:4}
\begin{aligned}
\mathbf{e}_n & =\mathbf{y}_n-\left[c_{\mathrm{desired}}, 0\right]^T, \\
\mathbf{a}_n & =\left[a_n, a_{n+1}, \ldots, a_{n+N-1}\right]^T \in \mathbb{R}^N .
\end{aligned}
\end{equation}

In Equation \ref{eq:2}, $P$ is a scalar used to amplify the terminal cost, thereby accelerating convergence to move the gripper until it reaches the desired state, where the $c_n$ converges to a desired value $c_{\mathrm{desired}}$ and  $v_n$ converges to zero. The control error, as defined in Equation \ref{eq:4}, incorporates weight coefficients $Q_c$, $Q_v$, and $Q_a$. Finally, an optimization problem can be formulated to compute the control inputs for the WSG gripper, enabling stable grasping during dynamic motions:
\begin{equation}
\label{eq:5}
\mathbf{a}_n^*=\arg \min J\left(\mathbf{y}_n, \mathbf{a}_n\right),
\end{equation}
subject to Equation \ref{eq:1} and the upper and the lower limits of $p_n$, $v_n$, and  $a_n$. The parameters for tactile-MPC on GelSight Mini are shown in Table \ref{table:mpcpara}.

\begin{table}[t]
\centering
\caption{Parameters for tactile-MPC controller using GelSight Mini.}
\begin{tabular}{l l l l l l l l l l}
\hline
% \Xhline{2\arrayrulewidth}
$c_{\mathrm{desired}}$ & $Q_a$ & $Q_c$ & $Q_v$ & $P$ & $N$ & $K_c$ & $\Delta t$ & $freq.$ \\ \hline
$5500 $   & $1$  & $1$  & $2$  & $10$ & $30$ & $50000$  & 1/60     & $60$ Hz  \\ \hline
\end{tabular}
\label{table:mpcpara}
\end{table}
\textbf{Rubbing using linear actuators:} another challenge is enhancing the adaptability of the rubbing motion for objects of different dimensions, requiring varying motion ranges of linear actuators. We propose a linear state feedback controller to regulate this range. After stabilizing the WSG gripper position $p_{\text {stable}}$ post-grasping, the linear actuator's movement range $L$ is linearly correlated with the gripping width: $L = k_p \cdot p_{\text {stable}} + b$, where $k_p$ and $b$ are tuned parameters. For objects smaller than $15\mathrm{~mm}$, the rotation servo retracts inward by a fixed angle to prevent them from falling.

\subsection{Scooping Manipulation and Precise Insertion}
This section presents the methodologies employed to enable the proposed gripper to grasp flat objects and complete precise insertion, exemplified by a credit card.

\textbf{Scooping manipulation:} given the necessity of a sharp fingertip for grasping flat objects, we have adapted the design concept in \cite{do2023densetact} by employing a replaceable fingernail made of PLA. Leveraging the five degrees of freedom, we devised a simplified and adaptable maneuver suitable for objects of varying thicknesses. Fig. \ref{fig:freebody} depicts the schematic of the grasping maneuver. From the free-body diagram when the card has been lifted with only three contact points with the ground the two fingers, the equivalent moment of external forces at the center of mass, $M_{\text{all}}$, can be expressed as:
\begin{figure}[t]
\centering
\includegraphics[width=0.47\textwidth]{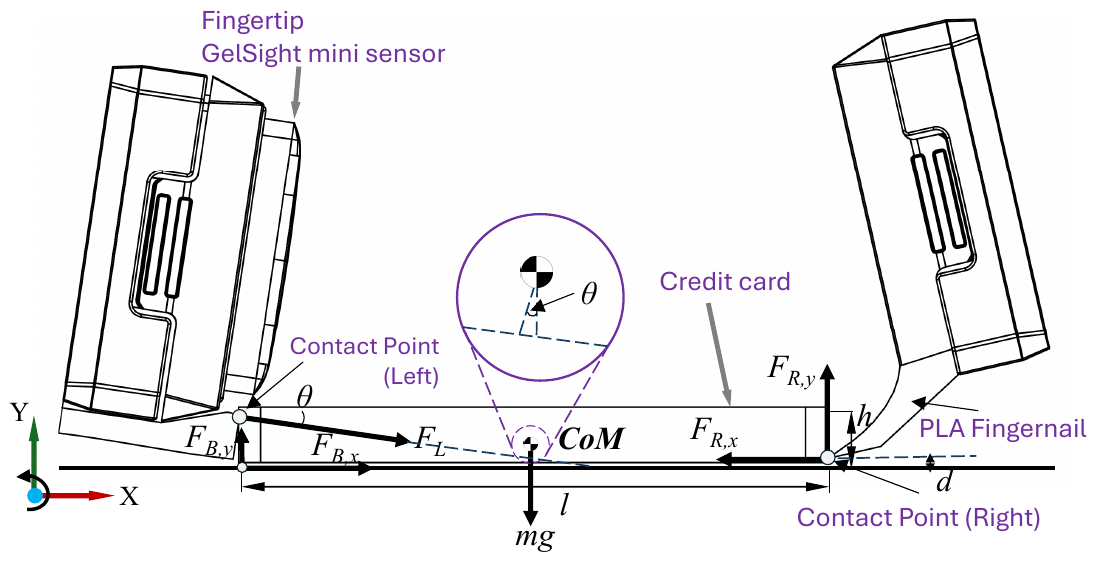}
\caption{Model for the designed scooping motion for the proposed gripper using a 3D printed fingernail}
\vspace{-2.0mm}
\label{fig:freebody}
\end{figure}
\begin{equation}
\label{eq:7}
\begin{aligned}
M_{\text{all}} &= \frac{1}{2}\left[-(h-2d)F_{R,x}+lF_{R,y}\right] \\
&\quad + \frac{1}{2}\left[hF_{B,x}-lF_{B,y}\right] + \frac{1}{2}\left[(l\tan\theta-h)F_L\cos\theta\right]
\end{aligned}
\end{equation}
where $F_{R,x}$ and $F_{R,y}$ denote the pressure force by the fingernail and the friction force by the ground. And $F_{B,x}$ and $F_{B,y}$ represent the friction force and the support force from the table. At the contact point of the left finger, the card experiences the positive force $F_L$. Additionally, $\mu_1$ and $\mu_2$ are the coefficients of friction for the two contacting surfaces and the object is subject to its gravitational force $mg$.

For the force analysis, we select a cross-section and make the following assumptions to simplify the model: within this cross-section, the contact between the finger nail and the finger is approximated as a point contact. Furthermore, we assume that the contact point on the object receives normal pressure from the vertical edges of the object and static frictional force along the direction of the object's surface.
\begin{equation}
\label{eq:8}
\begin{aligned}
F_{R,y} &= \mu_1 \cdot F_{R,x}, \\
F_{B,x} &= \mu_2 \cdot F_{B,y}, \\
F_{B,x}&= F_{R,x} - F_{L} \cdot \cos\theta , \\
F_{B,y} &= F_{L} \cdot \sin\theta - F_{R,y} + mg.
\end{aligned}
\end{equation}
\begin{equation}
\label{eq:9}
F_{R,x}(1 + \mu_1 \mu_2) = F_{L}(\mu_2 \sin\theta + \cos\theta) + \mu_2 mg
\end{equation}
\begin{equation}
\label{eq:10}
\begin{aligned}
M_{\text{all}} &= \frac{1}{2} F_{R,x} \left[ (2d - h + l\mu_1) \right] + \frac{1}{2} F_{B,y} \left[ h \mu_2 - l \right] \\
&\quad + \frac{1}{2} F_{L} \left[ (l \tan\theta - h) \cos\theta \right] \\
% &= \frac{1}{2} F_{R,X} \left[ 2d - h + 2l\mu_1 - \mu_1\mu_2h \right] \\
% &\quad + \frac{1}{2} F_{L} \left[ \sin\theta (h \mu_2 - l) +\cos\theta( l\tan\theta - h) \right] \\
% &\quad + \frac{1}{2} mg(h\mu_2 - l)\\
&= \frac{1}{2} F_{R,x} \left[ 2d - h + 2l\mu_1 - \mu_1\mu_2h \right] \\
&\quad + \frac{1}{2} \left[ \frac{1}{\mu_2 \sin\theta + \cos\theta} \left( F_{R,x}(1 + \mu_1 \mu_2) - \mu_2 mg \right) \right] \\
&\quad  \cdot \left[ \sin\theta (h \mu_2 - l) +\cos\theta( l\tan\theta - h) \right] \\
&= K_1 F_{R,x} + K_2.
\end{aligned}
\end{equation}

By considering the relationship between static friction and normal pressure, along with the force balance of the object, we derive Equation \ref{eq:8}, establishing the relationship between $F_{R,x}$ and $F_L$ as in Equation \ref{eq:9}. Solving these equations yields Equation \ref{eq:10}, where the friction coefficient, contact point height, length, thickness, and angle between $F_L$ and the ground are constant. To simplify, we introduce positive coefficients $K_1$ and $K_2$. Based on the direction of force $F_{R,x}$, the equivalent torque acts in the counterclockwise direction. According to our assumption, the direction of the equivalent torque remains constant throughout this process. Consequently, the object undergoes a counterclockwise flipping motion until the flipping action is complete.

% Our gripper design enables linear movement of the fingers, offering distinct advantages: the ability to adjust the position $d$ of the fingertip contact point and the angle $\theta$ between the finger and the tabletop. Within Equation \ref{eq:10}, increasing $d$ amplifies the coefficient $F_{R,X}$, thereby enhancing the moment for handling heavier objects. Conversely, adjusting $\theta$ modulates the coefficients $K_1$ and $K_2$, with $K_1$ decreasing and $K_2$ increasing as $\theta$ increases. Consequently, the linear degrees of freedom provide enhanced adaptability for handling objects with varying thicknesses and contribute to improved generalization capabilities.

\textbf{Precise grasping using tactile sensor:} after successfully grasping the card and securing it between the two fingers, we further exploit the advantages of tactile sensors to showcase the versatility of our gripper through an insertion task. This task involves two objectives: 1) using in-hand manipulation to adjust the orientation of the credit card to ensure that the chip is facing downward, and 2) achieving a precise grasping position ($x_d$, $y_d$). We formulate this problem as a model-based task, leveraging tactile information to estimate the geometric properties of the card.

Our action primitives consist of a linear pushing action by the manipulator and the grasping motion by the parallel DoF. The tactile-based exploration begins once the sensor establishes stable contact. Initially, with the long edge of the credit card in contact with the tabletop, our strategy for card manipulation involves sequentially moving the gripper in a 2D plane based on the numbers printed on the card from previous steps, and then release and adjusting the gripper's next grasp to align with the edge of the credit card corresponding to the digits. We set the exploration step along the long and short edge (or pushing action change) to be $s_x, s_y \in\{-2 \mathrm{~mm},-4 \mathrm{~mm},-8 \mathrm{~mm}\}$, depending on the current edge of the number on the card, as illustrated in Fig. 6. During the motion, the gripper is always perpendicular to the table. When the tactile image indicates that the last digit has reached a threshold position on the long edge of the credit card $x_d$, the robot arm ceases to push forward. To determine $y_d$ on the short edge, we first employ in-hand manipulation to rotate the card by adjusting the width of the parallel DoF. Given that the card is grasped at the far right end of the long edge, rotating around this point ensures that the card flips by the gravity, allowing the short edge of the credit card is in full contact with the tabletop. Subsequently, by iteratively performing the tactile exploration action, the gripper can eventually reach the desired gripping pose.

\begin{figure*}[!t]
  \centering
    {\includegraphics[width=1\textwidth]{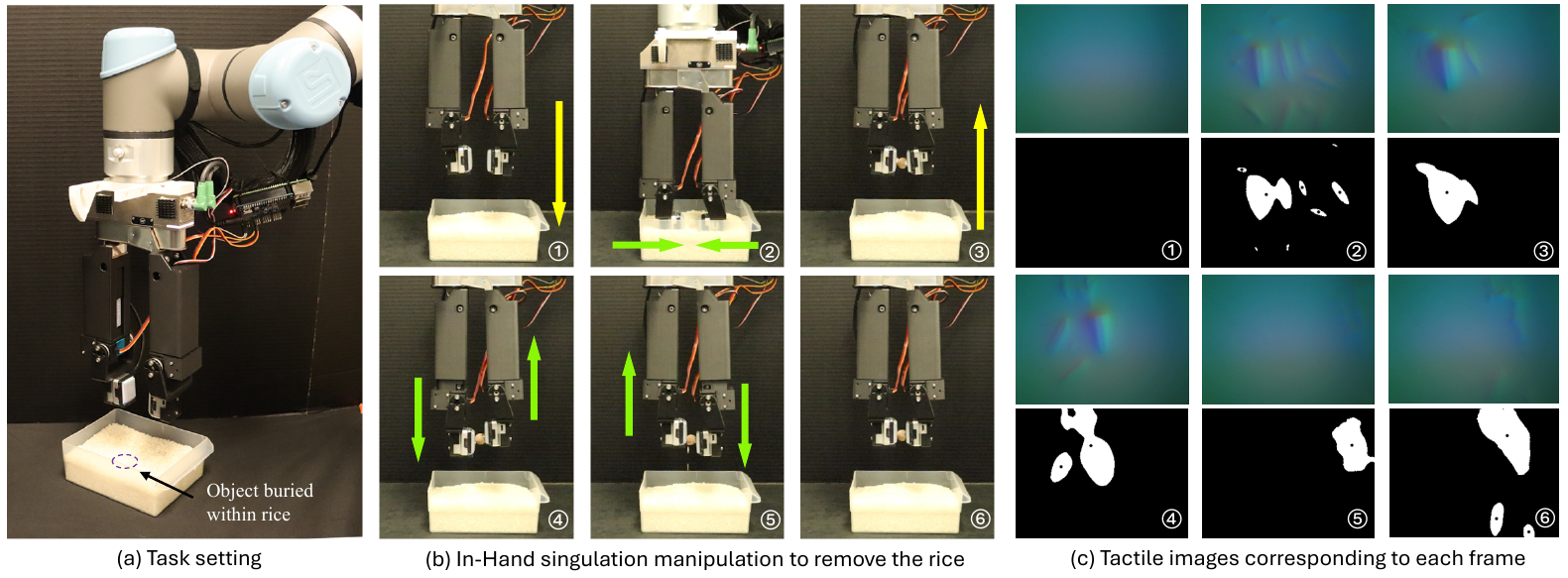}} \\
  \caption{Sequential images of in-hand grasping and singulation manipulation of a peanut buried in rice. The green arrow in (b) indicates the motion of the gripper, while the yellow arrow represents the motion of the robot arm. The same arrow notation is used in Figure \ref{fig:cardexp}(b).}
	\label{fig:singuexp}
\end{figure*}

\section{Experiment}
\subsection{Singulation and Classification}
\begin{figure}[t]
\centering
\includegraphics[width=0.48\textwidth]{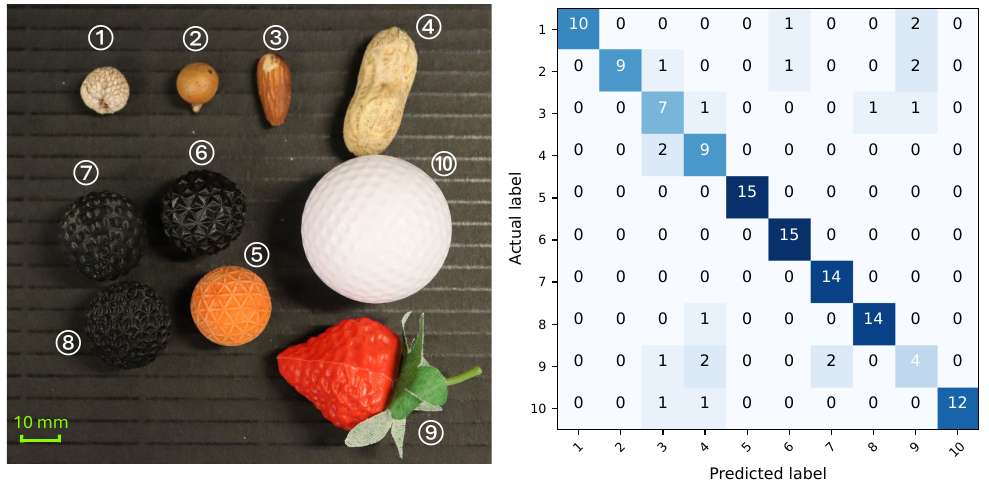}
\vspace{-5mm}
\caption{Selected objects for singulation task and the classification result.}
\label{fig:confusion}
\end{figure}

We employed a UR5e robot equipped with the proposed 5 DoF gripper. For this task, we assume that the initial position of all objects beneath the granular media remains constant, allowing the robot gripper to consistently grasp the object successfully. In the experimental setup, we selected rice as the granular media and placed it inside a plastic container measuring $135 \times 112 \times 55 \mathrm{~mm}$. The depth of rice filled in the container is approximately $30 \mathrm{~mm}$, as shown in Fig. \ref{fig:singuexp} (a). We selected a total of 10 objects with varying sizes, surface textures, geometric shapes, and physical properties, shown in Fig. \ref{fig:confusion}. These objects include small items (with diameters around $10\mathrm{~mm}$), two elliptical nuts, a deformable soft foam sphere, three 3D-printed rigid balls with PLA material featuring different surface textures, and two large objects, one golf ball with a diameter of $41\mathrm{~mm}$ and an artificial strawberry with an irregular geometric shape.

Fig. \ref{fig:singuexp}(b) and (c) depict sequential images and corresponding tactile images, including raw images and binary contact area images generated by thresholding the depth image for high precision. After the gripper is inserted into the rice, the tactile-MPC controller is engaged, enabling stable grasping by dynamically adjusting the grip width.Subsequently, in-hand manipulation is performed according to the proposed method to remove the rice trapped. As depicted in Fig. \ref{fig:singuexp}(c) corresponding to state $2$, due to residual material, the tactile image only shows the shape of rice grains, making classification processing difficult solely relying on tactile feedback. However, once the singulation manipulation is completed, state $6$ clearly depicts the features of the grasped peanut. 

The results of the singulation experiment (15 trials each) are presented in Table \ref{tab:sinresult}. The success rate is defined as the object remaining stable during the singulation process. For all objects combined, the overall success rate was 114 out of 150 attempts (76\%). Specifically, for different sphere-like objects, including balls and seeds, the success rate was notably high at 99 out of 105 attempts (\textbf{94.3\%}).

Despite most granular media being removed during the rubbing motion, minor residual may still remain. To further evaluate singulation performance, we conducted a classification of successful attempts in the first step using \cite{sandler2018mobilenetv2}, with results displayed in the confusion matrix in Fig. \ref{fig:confusion}. The left image annotates each object's respective labels. The results indicate high classification accuracy for most objects, suggesting successful removal of attached granular media and confirming the efficacy of our method.

\subsection{Credit Card Insertion}
\begin{figure*}[!t]
  \centering
    {\includegraphics[width=1\textwidth]{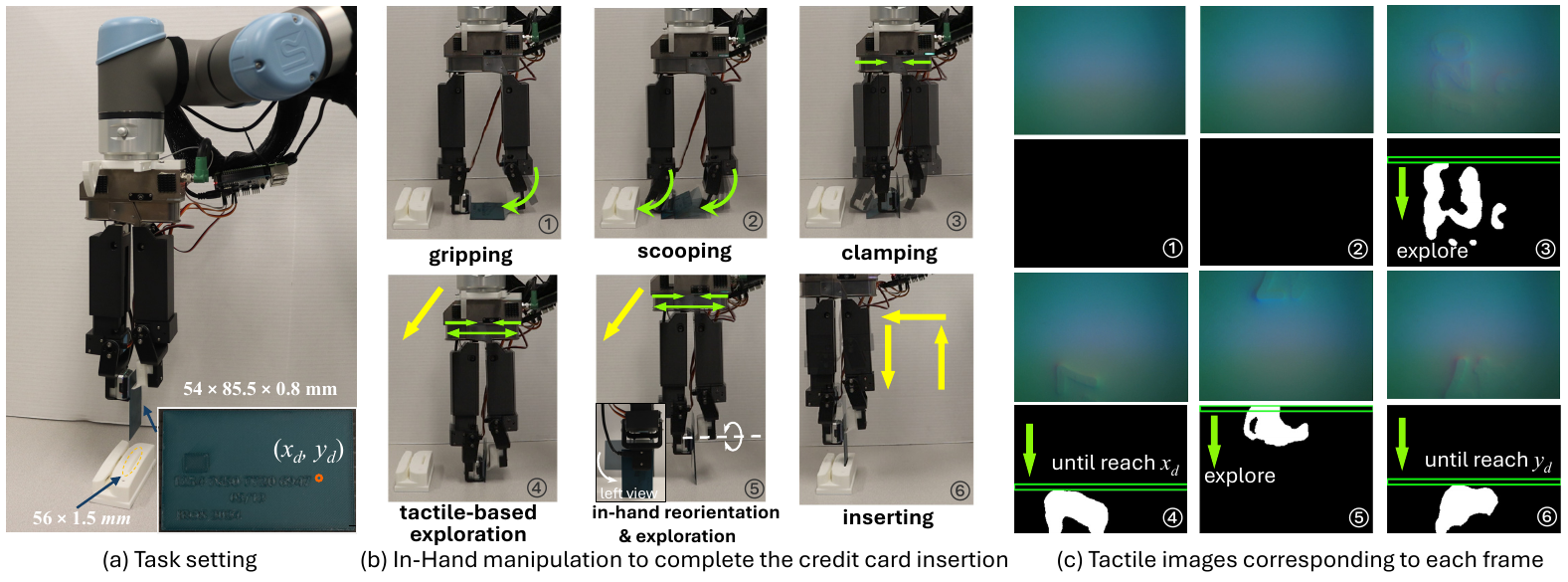}} \\
    \vspace{-1mm}
  \caption{Sequential images depicting in-hand scooping manipulation and precise insertion of a 3D printed credit card}
	\label{fig:cardexp}
 \vspace{-1mm}
\end{figure*}

We make the following assumptions to simplify the problem: 1) the pose of the card reader is predetermined, and 2) the initial position of the credit card is within a specified range beneath the gripper, ensuring that the gripper can always descend to grasp it. Fig. \ref{fig:cardexp}(a) shows the task setting. A 3D-printed credit card, measuring $85.5 \times 54 \times 0.8\mathrm{~mm}$ (or $1.2\mathrm{~mm}$ thick if considering raised numbers), was utilized. Additionally, a card reader for insertion was printed, featuring a hole with dimensions of $56 \times 1.5\mathrm{~mm}$.

\begin{table}[t]
    \caption{Singulation Experiment Results on 10 Selected Objects}
    \scriptsize  % Use \scriptsize for smaller font
    \centering
    \setlength{\tabcolsep}{1pt} % Adjust column separation for narrower table
    \begin{tabular}{c|cc|cc|c|ccc|cc}
    \toprule
    \multirow[b]{2}{*}{\textbf{Objects}} & \multicolumn{2}{c|}{\textbf{Small}} & \multicolumn{2}{c|}{\textbf{Nuts}} & \multicolumn{1}{c|}{\textbf{Soft}}  & \multicolumn{3}{c|}{\textbf{Rigid}} & \multicolumn{2}{c}{\textbf{Large}}\\
    & \makecell{Tree\\Seed 1} & \makecell{Tree\\Seed 2} & Almond & Peanut & \makecell{Soft\\Ball} & \makecell{PLA\\Ball 1} & \makecell{PLA\\Ball 2} & \makecell{PLA\\Ball 3} & \makecell{Artificial\\Strawberry} & \makecell{Golf\\Ball} \\
    \midrule
    success  & 13/15 & 13/15 & 10/15 & 11/15 & 15/15 & 15/15 & 14/15 & 15/15 & 9/15 & 14/15 \\
    \bottomrule
    \end{tabular}
    \label{tab:sinresult}
\end{table}

Fig. \ref{fig:cardexp}(b) and (c) depict the critical states and tactile feedback during the experiment. We directly employ the raw difference image and apply Canny detection followed by a series of morphological operations to extract the contact edge based on the numerical information for faster processing. In the initial scooping stage, the fingers' edges are parallel, leaving ample space between them to accommodate different card postures. This initial grasping step constrains the card's position, ensuring its stability relative to the fingers before each scooping action begins. During the scooping process, the GelSight fingers remain stationary, preventing the card from sliding and causing task failure, thus avoiding the complicated process of determining the appropriate angle. Upon completion of the flip, the tactile information of the card is transmitted to the GelSight. This denotes the credit card's orientation based on its numbers, distinguishing between its front and back sides. In the absence of tactile sensor feedback, the gripper executes an in-hand reorientation to flip the card, ensuring that the side with the numbers contacts the sensor. During exploration along the $x$-direction of the card, the robot arm continues to grasp and release until the upper contact edge (denoted by the green line) reaches $x_d$, as indicated in Fig. \ref{fig:cardexp}(c) from state $3$ to state $4$. Similarly, following the in-hand reorientation that adjusts the card to a vertical position, the robot arm proceeds with exploration along the $y$-direction until the desired pose $y_d$ is achieved. Finally, the gripper is lifted to complete the insertion. Our method achieved a \textbf{100\%} success rate for the complete task including scooping and insertion, in 10 consecutive trials with the credit card placed in random initial pose.

\section{Discussion and Conclusion}
Several issues observed in the experimental results merit discussion. In the singulation task, the success rate for non-spherical objects was notably lower. This disparity can be attributed to the variability in width during rotation, which is less stable compared to spherical objects. Sudden changes in width may lead to objects being dropped or squeezed out, and the use of a flat sensor pad exacerbates this effect. Developing a policy using a learning-based approach may offer a potential solution for improving performance in handling these objects, a direction we defer to future work.

In this paper, we introduce a five degrees of freedom tactile gripper capable of performing dexterous in-hand manipulation tasks. We evaluated the performance of our gripper and control methods through two dexterous tasks: singulation manipulation of $10$ various objects, and scooping and precision insertion of a credit card. The results underscored the efficacy of the gripper design and control strategies proposed. Overall, we achieved a 76\% success rate for the rubbing motion, with a success rate exceeding 94\% for sphere-shaped objects. Furthermore, the scooping manipulation exhibited a 100\% success rate. In future endeavors, we aim to expand our approach to tackle more intricate manipulation tasks.

\section*{Acknowledgment}
The authors would like to thank Zhengtong Xu for the invaluable discussions on the MPC controller and Guanlan Zhang for developing the early gripper prototype.
% \addtolength{\textheight}{-5cm}

\bibliographystyle{IEEEtran}
\bibliography{bibliography}

\end{document}